\documentclass[a4,11pt]{article}
\usepackage{acl2016}
\usepackage{times}
\usepackage{latexsym}
\usepackage[hyphens]{url}
\usepackage{paralist}
\aclfinalcopy 

\usepackage{graphicx}
\usepackage{linguex}
\usepackage{enumitem}
\usepackage{xcolor}
\usepackage{multirow}

\title{The LAMBADA dataset:\\Word prediction requiring a broad discourse context\Thanks{Denis and Germ\'an share first authorship. Marco, Gemma, and Raquel share senior authorship.}}

\author{Denis Paperno, Germ\'{a}n Kruszewski, Angeliki Lazaridou, Quan Ngoc Pham$^\dagger$, \\ 
	\textbf{Raffaella Bernardi, Sandro Pezzelle, Marco Baroni, Gemma Boleda, Raquel Fern\'{a}ndez}$^\ddagger$\\
	      CIMeC - Center for Mind/Brain Sciences, University of Trento\\
	      {\tt \{firstname.lastname\}@unitn.it}, $^{\dagger}${\tt quanpn90@gmail.com}\\[3pt]
     $^\ddagger$Institute for Logic, Language \& Computation, University of Amsterdam\\
	{\tt raquel.fernandez@uva.nl} }

\date{}

\begin{document}

\maketitle

\begin{abstract}
  We introduce LAMBADA, a dataset to evaluate the capabilities of
  computational models for text understanding
  by means of a word prediction task. LAMBADA is a collection of
  narrative passages sharing the characteristic that human subjects
  are able to guess their last word if they are exposed to the whole
  passage, but not if they only see the last sentence preceding the
  target word. To succeed on LAMBADA, computational models
  cannot simply rely on local context, but must be able to
  keep track of information in the broader discourse. We show that
  LAMBADA exemplifies a wide range of linguistic phenomena, and that
  none of several state-of-the-art language models reaches accuracy
  above 1\% on this novel benchmark. We thus propose LAMBADA as a challenging
  test set, meant to encourage the development of new models capable of genuine
  understanding of broad context in natural language text.
\end{abstract}

\section{Introduction}
\label{sec:intro}

The recent spurt of powerful end-to-end-trained neural networks for
Natural Language Processing
\cite[a.o.]{Hermann:etal:2015,Rocktaeschel:etal:2016,Weston:etal:2015b}~has
sparked interest in tasks to measure the progress they are bringing
about in genuine language understanding. Special care must be taken in
evaluating such systems, since their effectiveness at picking
statistical generalizations from large corpora can lead to the
illusion that they are reaching a deeper degree of understanding than
they really are. For example, the end-to-end system of
\newcite{Vinyals:Le:2015}, trained on large conversational datasets,
produces dialogues such as the following:

\ex.
\noindent{}\textbf{Human:} what is your job?\\
\noindent{}\textbf{Machine:} i'm a lawyer\\
\noindent{}\textbf{Human:} what do you do?\\
\noindent{}\textbf{Machine:} i'm a doctor

\noindent
Separately, the system responses are appropriate for the respective
questions. However, when taken together, they are incoherent. The
system behaviour is somewhat parrot-like. It can locally produce
perfectly sensible language fragments, but it fails to take
the meaning of the \emph{broader discourse context} into account. Much research
effort has consequently focused on designing systems able to keep
information from the broader context into memory, and possibly even
perform simple forms of reasoning about it
\cite[a.o.]{Hermann:etal:2015,Hochreiter:Schmidhuber:1997,Ji:etal:2015,Mikolov:etal:2015,Sordoni:etal:2015,Sukhbaatar:etal:2015,Wang:Cho:2015}.

In this paper, we introduce the LAMBADA dataset (\textbf{LA}nguage
\textbf{M}odeling \textbf{B}roadened to \textbf{A}ccount for
\textbf{D}iscourse \textbf{A}spects). LAMBADA proposes a word
prediction task where the target item is \emph{difficult to guess}
(for English speakers) when only the sentence in which it appears is
available, but becomes \emph{easy} when a broader context is
presented. Consider Example \ref{ex:miscarriage} in Figure
\ref{fig:examples-lambada}. The sentence \textit{Do you honestly think
  that I would want you to have a $\_\_\_$?} has a multitude of
possible continuations, but the broad context 
clearly
indicates that the missing word is \textit{miscarriage}.

LAMBADA casts language understanding in the classic word prediction
framework of language modeling. We can thus use it to test several
existing language modeling architectures, including systems
with capacity to hold longer-term contextual memories. In our
preliminary experiments, none of these models came even remotely
close to human performance, confirming that LAMBADA is a
challenging benchmark for research on automated models of natural
language understanding.

\setcounter{ExNo}{0}

\begin{figure*}[p] \small \centering
\hrule

\

\ex. \label{ex:miscarriage}
\begin{description}[nolistsep]
\item[{\rm \em Context:}] ~``Yes, I thought I was going to lose the baby.''  ``I was scared too,''
 he stated, sincerity flooding his eyes.  ``You were ?'' ``Yes, of
 course.  Why do you even ask?'' ``This baby wasn't exactly planned
 for.''
 \item[\rm \em Target sentence:]~ ``Do you honestly think that I would want you to have a \_\_\_\_\_ ?''
 \item[\rm \em Target word:]~ miscarriage 
\end{description}

\hrule



\ex. \label{ex:Gabriel}
\begin{description}[nolistsep]
\item[{\rm \em Context:}]~ ``Why?"  ``I would have thought you'd find him rather dry," she said.  ``I don't know about that," said \underline{Gabriel}.  ``He was a great craftsman," said Heather.  ``That he was," said Flannery. 
\item[\rm \em Target sentence:]~ ``And Polish, to boot," said \_\_\_\_\_. 
\item[\rm \em Target word:]~Gabriel 
\end{description}

\hrule

\ex. \label{ex:chains} 
\begin{description}[nolistsep]
\item[{\rm \em Context:}]~ Preston had been the last person to wear those \underline{chains}, and I knew what I'd see and feel if they were slipped onto my skin-the Reaper's unending hatred of me.  I'd felt enough of that emotion already in the amphitheater.  I didn't want to feel anymore.  ``Don't put those on me,'' I whispered.  ``Please.''
\item[\rm \em Target sentence:]~ Sergei looked at me, surprised by my low, raspy please, but he put down the \_\_\_\_\_. 
\item[\rm \em Target word:]~ chains 
\end{description}

\hrule

\ex. \label{ex:dancing} 
\begin{description}[nolistsep]
\item[{\rm \em Context:}]~ They tuned, discussed for a moment, then struck up a lively jig.  Everyone joined in, turning the courtyard into an even more chaotic scene, people now \underline{dancing} in circles, swinging and spinning in circles, everyone making up their own dance steps.  I felt my feet tapping, my body wanting to move. 
\item[\rm \em Target sentence:] ~ Aside from writing, I 've always loved \_\_\_\_\_. 
\item[\rm \em Target word:]~ dancing
\end{description}

\hrule

\ex. \label{ex:camera} 
\begin{description}[nolistsep]
\item[{\rm \em Context:}]~ He shook his head, took a step back and
 held his hands up as he tried to smile without losing a cigarette.
 ``Yes you can,'' Julia said in a reassuring voice.  ``I 've already
 focused on my friend.  You just have to click the shutter, on top,
 here.''
\item[\rm \em Target sentence:]~ He nodded sheepishly, through his cigarette away and took the \_\_\_\_\_. 
\item[\rm \em Target word:]~ camera 
\end{description}

\hrule

\ex. \label{ex:signs} 
\begin{description}[nolistsep]
\item[{\rm \em Context:}]~ In my palm is a clear stone, and inside it is a small ivory statuette.  A guardian angel. ``Figured if you're going to be out at night getting hit by cars, you might as well have some backup.''  I look at him, feeling stunned.  Like this is some sort of \underline{sign}.
\item[\rm \em Target sentence:]~ But as I stare at Harlin, his mouth curved in a confident grin, I don't care about \_\_\_\_\_.
\item[\rm \em Target word:]~ signs 
\end{description}

\hrule

\ex. \label{ex:coffee}
\begin{description}[nolistsep]
\item[{\rm \em Context:}]~ Both its sun-speckled shade and the cool grass beneath were a welcome
respite after the stifling kitchen, and I was glad to relax against
the tree's rough, brittle bark and begin my breakfast of buttery,
toasted bread and fresh fruit.  Even the water was tasty, it was so clean and cold. 
\item[\rm \em Target sentence:]~ It almost made up for the lack of \_\_\_\_\_. 
\item[\rm \em Target word:]~ coffee
\end{description}

\hrule

\ex. \label{ex:lonely}
\begin{description}[nolistsep] 
\item[{\rm \em Context:}]~ My wife refused to allow me to come to Hong Kong when the plague was at its height and --" ``Your wife, Johanne?  You are married at last ?"  Johanne grinned.  ``Well, when a man gets to my age, he starts to need a few home comforts. 
\item[\rm \em Target sentence:] ~  After my dear mother passed away
 ten years ago now, I became \_\_\_\_\_.
\item[\rm \em Target word:]~ lonely
\end{description}

\hrule

\ex. \label{ex:died}
\begin{description}[nolistsep] 
\item[{\rm \em Context:}]~ ``Again, he left that up to you.  However, he was adamant in his
 desire that it remain a private ceremony.  He asked me to make sure,
 for instance, that no information be given to the newspaper regarding
 his death, not even an obituary.
\item[\rm \em Target sentence:] ~  
 I got the sense that he didn't want anyone, aside from the three of
 us, to know that he'd even \_\_\_\_\_.
\item[\rm \em Target word:]~ died 
\end{description}

\hrule

\ex. \label{ex:driving}
\begin{description}[nolistsep]
\item[{\rm \em Context:}] ~ The battery on Logan's radio must have been on the way out.  So he told himself.  There was no other explanation beyond Cygan and the staff at the White House having been overrun.  Lizzie opened her eyes with a flutter.  They had been on the icy road for an hour without incident. 
\item[\rm \em Target sentence:] ~ Jack was happy to do all of the \_\_\_\_\_. 
\item[\rm \em Target word:]~ driving 
\end{description}

\hrule

 \caption{Examples of LAMBADA passages. Underlined words highlight when the target word (or its lemma) occurs in the context. }
 \label{fig:examples-lambada}
\end{figure*}

\setcounter{ExNo}{1}

\section{Related datasets}
\label{sec:related-work}

The CNN/Daily Mail (CNNDM) benchmark recently introduced by
\newcite{Hermann:etal:2015} is closely related to LAMBADA.  CNNDM
includes a large set of online articles that are published together
with short summaries of their main points. The task is to guess a
named entity that has been removed from one such summary. Although the
data are not normed by subjects, it is unlikely that the missing named
entity can be guessed from the short summary alone, and thus, like in
LAMBADA, models need to look at the broader context (the
article).  Differences between the two datasets include text genres
(news vs.~novels; see Section~\ref{sec:data-coll-meth}) and the fact
that missing items in CNNDM are limited to named entities. Most
importantly, the two datasets require models to perform different
kinds of inferences over broader passages. For CNNDM, models must be
able to \emph{summarize} the articles, in order to make sense of the
sentence containing the missing word, whereas in LAMBADA the last
sentence is not a summary of the broader passage, but a
\emph{continuation} of the same story. Thus, in order to succeed,
models must instead understand what is a plausible development of a
narrative fragment or a dialogue.

Another related benchmark, CBT, has been introduced by
\newcite{Hill:etal:2015}. Like LAMBADA, CBT is a collection of book
excerpts, with one word randomly removed from the last sentence in a
sequence of 21 sentences. While there are other design differences,
the crucial distinction between CBT and LAMBADA is that the CBT
passages were not filtered to be human-guessable in the broader
context only. Indeed, according to the post-hoc analysis of a sample
of CBT passages reported by Hill and colleagues, in a large proportion
of cases in which annotators could guess the missing word from the
broader context, they could also guess it from the last sentence
alone. At the same time, in about one fifth of the cases, the
annotators could not guess the word even when the broader context was
given. Thus, only a small portion of the CBT passages are really
probing the model's ability to understand the broader context, which is
instead the focus of LAMBADA.

The idea of a book excerpt completion task was originally introduced
in the MSRCC dataset \cite{Zweig:Burges:2011}. However, the latter
limited context to single sentences, not attempting to measure
broader passage understanding.

Of course, text understanding can be tested through other tasks,
including entailment detection \cite{Bowman:etal:2015}, answering
questions about a text \cite{Richardson:etal:2013,Weston:etal:2015b}
and measuring inter-clause coherence \cite{Yin:Schuetze:2015}. While
different tasks can provide complementary insights into the models'
abilities, we find word prediction particularly attractive because of
its naturalness (it's easy to norm the data with non-expert humans)
and simplicity. Models just need to be trained to predict the most
likely word given the previous context, following the classic
language modeling paradigm, which is a much simpler setup than the one
required, say, to determine whether two sentences entail each
other. Moreover, models can have access to virtually unlimited amounts
of training data, as all that is required to train a language model is
raw text.  On a more general methodological level, word prediction has
the potential to probe almost any aspect of text understanding,
including but not limited to traditional narrower tasks such as
entailment, co-reference resolution or word sense disambiguation.


\section{The LAMBADA dataset}
\label{sec:the-dataset}

\subsection{Data collection\footnote{Further technical details are provided in the Supplementary Material (SM): \url{http://clic.cimec.unitn.it/lambada/}}}
\label{sec:data-coll-meth}

LAMBADA consists of \emph{passages} composed of a \emph{context} (on average 4.6 sentences) and a
\emph{target sentence}. The context size is the minimum number of complete sentences before the target sentence such that they cumulatively contain at least 50 tokens (this size was chosen in a pilot study). The task is to guess the last word of the target 
sentence (the \emph{target word}). The constraint that the target word be
the last word of the sentence, while not necessary for our research
goal, makes the task more natural for human subjects.

The LAMBADA data come from the Book Corpus \cite{zhu15}. The fact that it contains
unpublished novels minimizes the potential usefulness of
general world knowledge and external resources for the task, in contrast to other kinds of texts like news data, Wikipedia text, or famous
novels. The corpus, after duplicate removal and filtering out of
potentially offensive material with a stop word list, %
contains 5,325 novels and 465 million words. We
randomly divided the novels into equally-sized training and
development+testing partitions. We built the LAMBADA dataset from the
latter, with the idea that models tackling LAMBADA should be trained
on raw text from the training partition, composed of 2662 novels and encompassing more than
200M words. Because novels are
pre-assigned to one of the two partitions only, LAMBADA passages
 are self-contained and cannot be solved by exploiting the knowledge in the remainder of the
novels, for example background information about the characters involved or the properties of the fictional world in a given novel. The same novel-based division method is used
to further split LAMBADA data between development and testing.

To reduce time and cost of dataset collection, we filtered out
passages that are relatively easy for standard language models, since
such cases are likely to be guessable based on local context alone. We
used a combination of four language models, chosen by availability
and/or ease of training: a pre-trained recurrent neural network (RNN)
\cite{export:175562} and three models trained on the Book Corpus (a
standard 4-gram model, a RNN and a feed-forward
model; see SM for details, and note that these are different from the
models we evaluated on LAMBADA as described in Section
\ref{sec:baseline-methods} below). Any passage whose target word had
probability $\ge$0.00175 according to any of the language models was
excluded.



A random sample of the remaining passages were then evaluated by
human subjects %
through the CrowdFlower crowdsourcing
service\footnote{\url{http://www.crowdflower.com}} 
 in three steps. For a given passage,

\begin{enumerate}[nolistsep]
\item one human subject guessed the target word based on the whole passage (comprising the context and the target sentence); if the
 guess was right,
\item a second subject guessed the target word based on the whole passage; if that guess was also right, 
\item more subjects tried to guess the target word based on the target
  sentence only, until the word was guessed or the number of
  unsuccessful guesses reached 10; if no subject was able to guess the
  target word, the passage was added to the LAMBADA dataset.
\end{enumerate}

The subjects in step 3 were allowed 3 guesses per sentence, to
maximize the chances of catching cases where the target words were
guessable from the sentence alone. Step 2 was added based on a pilot
study that revealed that, while step 3 was enough to ensure that the
data could not be guessed with the local context only, step 1 alone
did not ensure that the data were easy given the discourse context
(its output includes a mix of cases ranging from obvious to relatively
difficult, guessed by an especially able or lucky step-1 subject).
We made sure that it was not possible for the same subject
to judge the same item in both passage and sentence conditions
(details in SM).

In the crowdsourcing pipeline,  84--86\% items were discarded at step 1, an additional 6--7\% at step 2 and another 3--5\% at step 3. Only about one in 25 input examples passed all the selection steps.

Subjects were paid \$0.22 per page in steps 1 and 2 (with 10 passages
per page) and \$0.15 per page in step 3 (with 20 sentences per
page). Overall, each item in the resulting dataset costed \$1.24 on
average.  Alternative designs, such as having step 3 before step 2 or
before step 1, were found to be more expensive. Cost considerations also precluded us from using more subjects at stage 1, which could in principle improve the quality of filtering at this step.

Note that the criteria for passage inclusion were very strict: We
required two consecutive subjects to exactly match the missing word,
and we made sure that no subject (out of ten) was able to provide it
based on local context only, even when given 3 guesses.  An
alternative to this perfect-match approach would have been to include
passages where broad-context subjects provided other plausible or synonymous
continuations.  However, it is very challenging, both practically and methodologically, to
determine which answers other than the original fit the passage well,
especially when the goal is to distinguish between items that are
solvable in broad-discourse context and those where the local context
is enough. Theoretically, substitutability in context could be tested with manual annotation by multiple additional raters, but this would not be financially or practically feasible for a dataset of this scale (human annotators received over 200,000 passages at stage 1). For this reason we went for the strict hit-or-miss approach,
keeping only items that can be unambiguously determined by human
subjects.


\subsection{Dataset statistics}
\label{sec:dataset-statistics}

The LAMBADA dataset consists of 10,022 passages, divided into 4,869
development and 5,153 test passages (extracted from 1,331 and 1,332
disjoint novels, respectively). The average passage consists of 4.6 sentences in the context plus 1 target sentence, for a total length of 75.4 tokens (dev)
/ 75 tokens (test). Examples of
passages in the dataset are given in
Figure~\ref{fig:examples-lambada}.

The training data for language models to be tested on LAMBADA include
the full text of 2,662 novels (disjoint from those in dev+test),
comprising 203 million words.  
Note that the training data consists of text from the same domain as
the dev+test passages, in large amounts but not filtered in the same
way. This is partially motivated by economic considerations (recall
that each data point costs \$1.24 on average), but, more importantly,
it is justified by the intended use of LAMBADA as a tool to evaluate
general-purpose models in terms of how they fare on broad-context
understanding (just like our subjects could predict the missing words
using their more general text understanding abilities), not as a
resource to develop \emph{ad-hoc} models only meant to predict the
final word in the sort of passages encountered in LAMBADA. The
development data can be used to fine-tune models to the specifics of
the LAMBADA passages.

\begin{figure*}[ht]
\begin{tabular}{@{\hspace*{-.3cm}}c@{}c@{}c@{}}
\includegraphics[height=4.8cm]{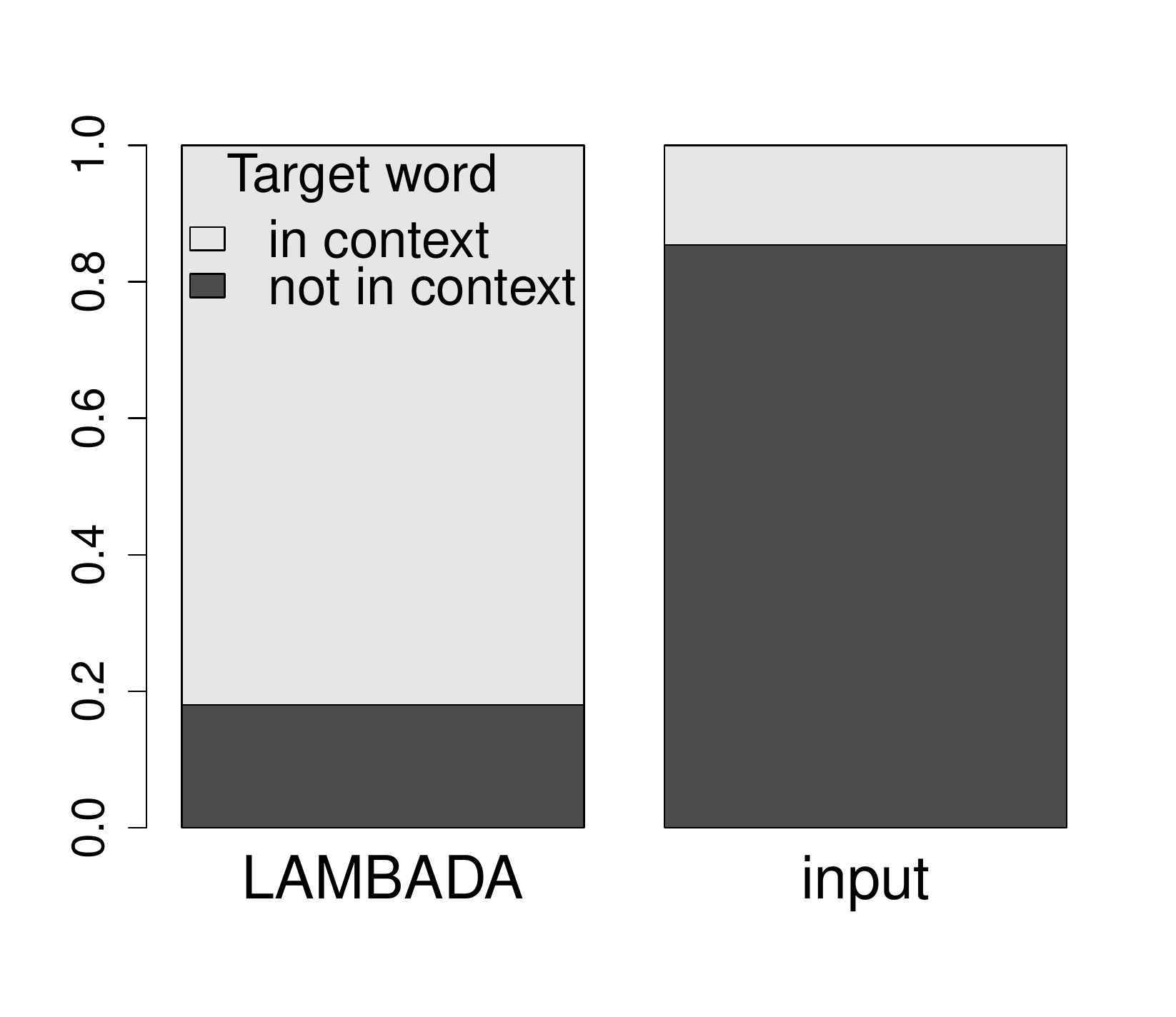}&
\includegraphics[height=5cm]{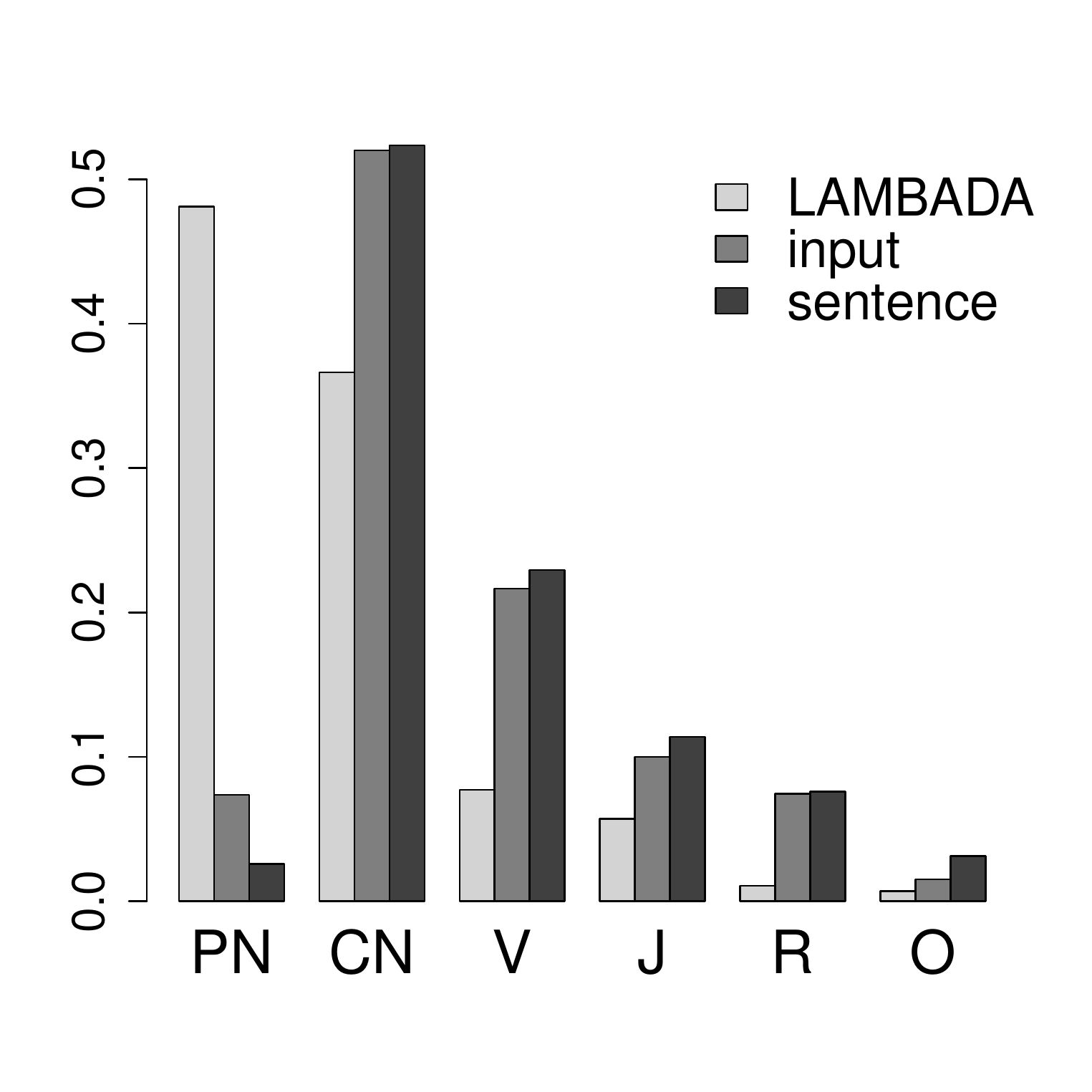}&
\includegraphics[height=5cm]{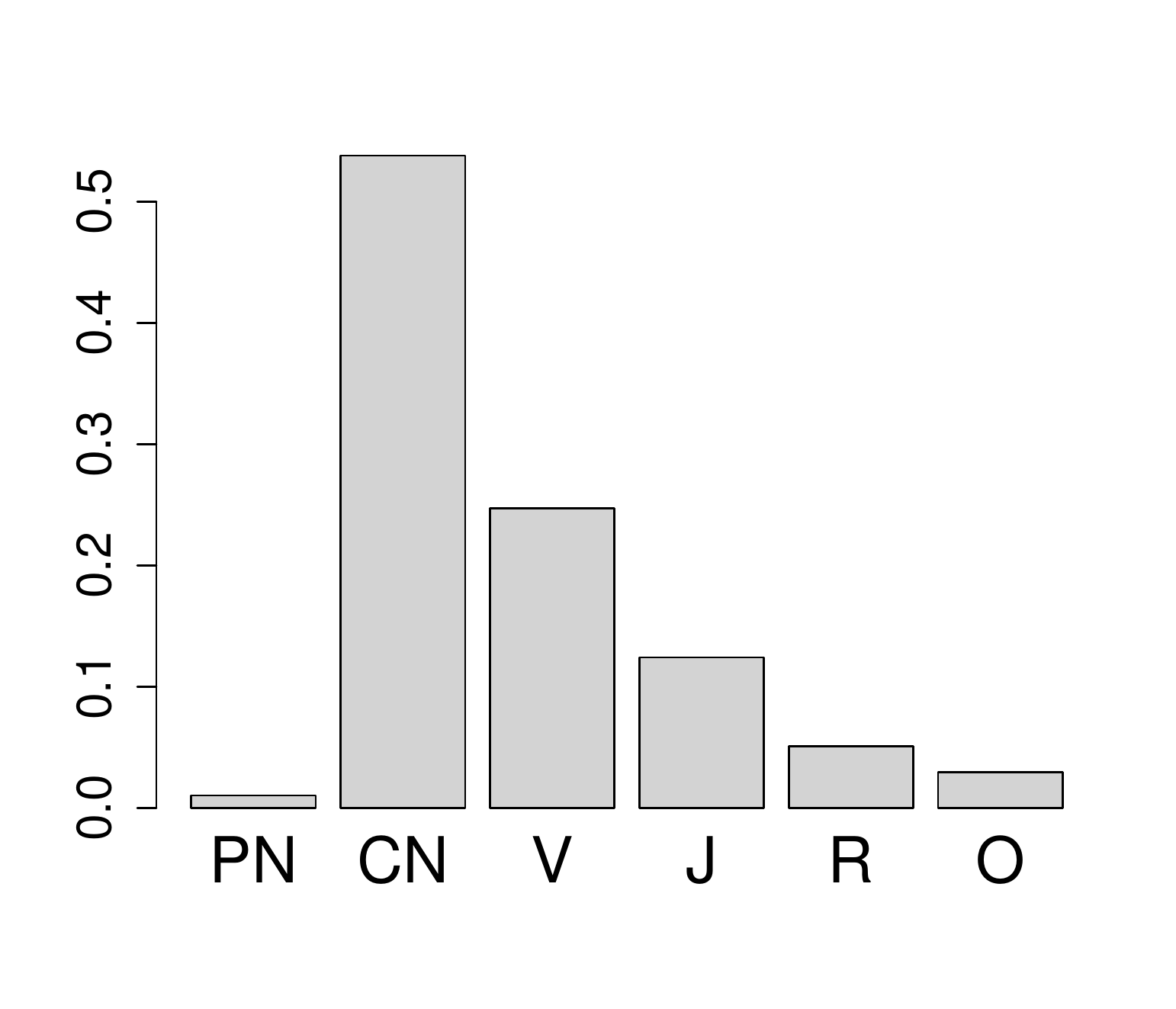}\\[-10pt]
(a) & (b) & (c)
\end{tabular}
\caption{(a) Target word in or not in context; (b) Target word POS
  distribution in LAMBADA vs.\ data presented to human
  subjects (input) and items guessed with sentence context only
  (PN=proper noun, CN=common noun, V=verb, J=adjective,
  R=adverb, O=other); (c)
  Target word POS distribution of LAMBADA passages where the lemma of
  the target word is not in the context (categories as in (b)).}
\label{fig:plots}
\end{figure*}

\subsection{Dataset analysis}
\label{sec:analysis}

Our analysis of the LAMBADA data suggests that, in order for the
target word to be predictable in a broad context only, it must be
strongly cued in the broader discourse. Indeed, it is typical for
LAMBADA items that the target word (or its lemma) occurs in the
context. Figure~\ref{fig:plots}(a) compares the LAMBADA items to 
a random 5000-item sample from the input data,
that is, the passages that
were presented to human subjects in the filtering phase (we sampled 
from all passages passing the automated filters described in Section
\ref{sec:data-coll-meth} above,  including those that made it to LAMBADA). 
The figure shows that when subjects guessed the word (only) in the
broad context, often the word itself occurred in the context: More
than 80\% of LAMBADA passages include the target word in the context,
while in the input data that was the case for less than 15\% of the
passages. To guess the right word, however, subjects must still put
their linguistic and general cognitive skills to good use, as shown by
the examples featuring the target word in the context reported in
Figure~\ref{fig:examples-lambada}.


Figure~\ref{fig:plots}(b) shows that most target words in LAMBADA are
proper nouns (48\%), followed by common nouns (37\%) and, at a
distance, verbs (7.7\%). In fact, proper nouns are hugely
over-represented in LAMBADA, while the other categories are
under-represented, compared to the POS distribution in the input.
A variety of factors converges in making proper nouns easy
for subjects in the LAMBADA task. In particular, when the context
clearly demands a referential expression, the constraint that the
blank be filled by a single word excludes other possibilities such as
noun phrases with articles, and there are 
reasons to suspect that
co-reference is easier than other discourse phenomena in our task (see
below). However, although co-reference seems to play a big role, 
only 0.3\% of target words are pronouns.

%

Common nouns are still pretty frequent in LAMBADA, constituting over one
third of the data. Qualitative analysis reveals a mixture of
phenomena. Co-reference is again quite common (see
Example~\ref{ex:chains} in Figure \ref{fig:examples-lambada}),
sometimes as ``partial'' co-reference facilitated by
bridging mechanisms ({\em shutter--camera}; Example~\ref{ex:camera})
or through the presence of a near synonym ({\em`lose the
  baby'--miscarriage}; Example \ref{ex:miscarriage}). However, we also
often find other phenomena, such as
the inference of prototypical participants
in an event. For instance, if the passage describes someone having
breakfast together with typical food and beverages (see Example
\ref{ex:coffee}), subjects can
guess the target word \emph{coffee} without it having been explicitly
mentioned.

In contrast, verbs, adjectives, and adverbs are rare in LAMBADA. Many
of those items can be guessed with local sentence context only,
as shown in Figure~\ref{fig:plots}(b), which  also reports the POS
distribution of the set of items that
were guessed by subjects based on the target-sentence context only
(step 3 in Section \ref{sec:data-coll-meth}). Note a
higher proportion of verbs, adjectives and adverbs in the latter
set in Figure~\ref{fig:plots}(b). 
While end-of-sentence context skews input distribution in favour of nouns, 
subject filtering does show a clear differential effect for nouns vs.~other POSs. 
 Manual inspection reveals that broad context is not necessary to guess items like frequent verbs ({\em ask, answer, call}), adjectives, and closed-class adverbs ({\em now}, {\em too}, {\em well}),
as well as time-related adverbs ({\em quickly}, {\em recently}).  In
these cases, the sentence context suffices, so few of them end
up in LAMBADA (although of course there are exceptions, such as
Example~\ref{ex:lonely}, where
the target word is an adjective).  This contrasts with other types of
open-class adverbs (e.g., {\em innocently, confidently}), which are
generally hard to guess with both local and broad context. %
The low proportion of these kinds of adverbs and of verbs among guessed
items in general suggests that tracking event-related phenomena (such
as script-like sequences of events) is harder for subjects than
co-referential phenomena, at least as framed in the LAMBADA task.
Further research is needed to probe this hypothesis.

Furthermore, we observe that, while explicit mention in the preceding
discourse context is critical for proper nouns, the other categories
can often be guessed without having been explicitly introduced. This
is shown in Figure~\ref{fig:plots}(c), which depicts the POS
distribution of LAMBADA items for which the lemma of the target word
is not in the context (corresponding to about 16\% of LAMBADA in
total).\footnote{The apparent 1\% of out-of-context proper nouns shown
  in Figure~\ref{fig:plots}(c) is due to lemmatization mistakes
  (fictional characters for which the lemmatizer did not recognize a
  link between singular and plural forms, e.g., {\em Wynn -- Wynns}). A
  manual check confirmed that all proper noun target words in LAMBADA
  are indeed also present in the context.} %
Qualitative analysis of items with verbs and adjectives as targets
suggests that the target word, although not present in
the passage, is still strongly implied by the context. In about one
third of the cases examined, the missing word is ``almost there''.
For instance, the passage contains a word with the same root but a
different part of speech (e.g., {\em death--died} in
Example~\ref{ex:signs}), or a synonymous expression (as mentioned
above for ``miscarriage''; we find the same phenomenon for verbs,
e.g., {\em `deprived you of water'--dehydrated}).

In other cases, correct prediction requires more complex discourse
inference, including guessing prototypical participants of a scene
(as in the \emph{coffee} example above), actions or events
strongly suggested by the discourse (see Examples~\ref{ex:miscarriage}
and \ref{ex:driving}, where the mention of an icy road helps in
predicting the target {\em driving}), or qualitative properties of
participants or situations (see Example~\ref{ex:lonely}). Of course,
the same kind of discourse reasoning takes place when the target word
is already present in the context (cf.~Examples~\ref{ex:chains} and \ref{ex:dancing}). The presence of the
word in context does not make the reasoning unnecessary (the task
remains challenging), but facilitates the inference.

As a final observation, intriguingly, the LAMBADA items contain
(quoted) direct speech significantly more often than the input items
overall (71\% of LAMBADA items vs.~61\% of items in the input sample),
see, e.g., Examples \ref{ex:miscarriage} and \ref{ex:Gabriel}. Further
analysis is needed to investigate in what way more dialogic discourse
might facilitate the prediction of the final target word.

In sum, LAMBADA contains a myriad of
phenomena that, besides making it challenging from the text
understanding perspective, are of great interest to the broad Computational
Linguistics community. %
To return to Example~\ref{ex:miscarriage}, solving it requires a
combination of linguistic skills ranging from (morpho)phonology (the plausible
target word \emph{abortion} is ruled out by the indefinite determiner
\emph{a}) through morphosyntax (the slot should be filled by a common
singular noun) to pragmatics (understanding what the male participant
is inferring from the female participant's words), in addition to
general reasoning skills. It is not surprising, thus, that LAMBADA is
so challenging for current models, as we show next.

\section{Modeling experiments}
\label{sec:baseline-methods}


\paragraph{Computational methods} We tested several existing language models and baselines on
LAMBADA. We implemented a simple \textbf{RNN} \cite{Elman:1990}, a
Long Short-Term Memory network (\textbf{LSTM})
\cite{Hochreiter:Schmidhuber:1997}, a traditional statistical
\textbf{N-Gram} language model \cite{Stolcke:2002} with and without
\textbf{cache}, and a \textbf{Memory Network}
\cite{Sukhbaatar:etal:2015}. We remark that at least LSTM, Memory
Network and, to a certain extent, the cache N-Gram model have, among
their supposed benefits, the ability to take broader contexts into
account. Note moreover that variants of RNNs and LSTMs are at the
state of the art when tested on standard language modeling benchmarks
\cite{Mikolov:2014}. Our Memory Network implementation is similar to
the one with which \newcite{Hill:etal:2015} reached the best results
on the CBT data set (see Section \ref{sec:related-work} above). While
we could not re-implement the models that performed best on CNNDM (see
again Section \ref{sec:related-work}), our LSTM is architecturally
similar to the Deep LSTM Reader of \newcite{Hermann:etal:2015}, which
achieved respectable performance on that data set. Most importantly,
we will show below that most of our models reach impressive
performance when tested on a more standard language modeling data set
sourced from the same corpus used to build LAMBADA. This
\textbf{control} set was constructed by randomly sampling 5K passages
of the same shape and size as the ones used to build LAMBADA from the
same test novels, but without filtering them in any way. Based
on the control set results, to be discussed below, we can reasonably
claim that the models we are testing on LAMBADA are very good at
standard language modeling, and their low performance on the latter
cannot be attributed to poor quality.

In order to test for strong biases in the data, we constructed
\textbf{Sup-CBOW}, a baseline model weakly tailored to the task at
hand, consisting of a simple neural network that takes as input a
bag-of-word representation of the passage and attempts to predict the
final word. The input representation comes from adding pre-trained
CBOW vectors \cite{Mikolov:etal:2013b} of the words in the
passage.\footnote{\url{http://clic.cimec.unitn.it/composes/semantic-vectors.html}}
We also considered an unsupervised variant (\textbf{Unsup-CBOW}) where
the target word is predicted by cosine similarity between the passage
vector and the target word vector. Finally, we evaluated several
variations of a random guessing baseline differing in terms of the
word pool to sample from. The guessed word could be picked from: the
full vocabulary, the words that appear in the current passage and a
random uppercased word from the passage. The latter baseline aims at
exploiting the potential bias that proper names account for a
consistent portion of the LAMBADA data (see Figure \ref{fig:plots}
above).

Note that LAMBADA was designed to challenge language models with
harder-than-average examples where broad context understanding is
crucial. However, the average case should not be disregarded either, since we
want language models to be able to handle both cases. For this
reason, we trained the models entirely on unsupervised data and expect
future work to follow similar principles. Concretely, we trained the
models, as is standard practice, on predicting each upcoming word
given the previous context, using the LAMBADA training data (see
Section \ref{sec:dataset-statistics} above) as input corpus.  The only exception to this procedure was Sup-CBOW where
we extracted from the training novels similar-shaped passages to those
in LAMBADA and trained the model on them (about 9M
passages). Again, the goal of this model was only to test for
potential biases in the data and not to provide a full account for the
phenomena we are testing.  We restricted the vocabulary of the models
to the 60K most frequent words in the training set (covering $95\%$ of
the target words in the development set). %
The model hyperparameters were tuned on their accuracy in the
development set. The same trained models were tested on the LAMBADA
and the control sets. See SM for the tuning
details. 

\newcommand{\expnumber}[2]{{#1}\mathrm{e}{#2}}

\begin{table*}[htb]
  \centering
	
  \begin{tabular}{|l|l|c|c|c|}
    \hline
    Data & Method & Acc. & Ppl. & Rank\\
    \hline
    \multirow{10}{0.7in}{LAMBADA} 
    &\multicolumn{4}{c|}{\emph{baselines}}\\
    \cline{2-5}
    &  Random vocabulary word & $0$ & $60000$ & $30026$\\
    &  Random word from passage & $1.6$ & - & -\\
    &  Random capitalized word from passage & $7.3$ & - & -\\
    & Unsup-CBOW & $0$ & $57040$ & $16352$ \\
    & Sup-CBOW & $0$ & $47587$ & $4660$\\
    \cline{2-5}
    &\multicolumn{4}{c|}{\emph{models}}\\
    \cline{2-5}
    & N-Gram & $0.1$ & $3125$ & $993$ \\
    & N-Gram w/cache & $0.1$ & $768$ & $87$ \\
    & RNN & $0$ &$14725$ &$7831$\\
    & LSTM & $0$ & $5357$ & $324$ \\
    & Memory Network & $0$ & $16318$ & $846$ \\
    \hline
    \multirow{10}{0.7in}{Control} 
    &\multicolumn{4}{c|}{\emph{baselines}}\\
    \cline{2-5}
    &  Random vocabulary word & $0$ & $60000$ & $30453$\\
    &  Random word from passage & $0$ &  - & -\\
    &  Random capitalized word from passage & $0$ & - & -\\
    & Unsup-CBOW & $0$ & $55190$ & $12950$ \\
    & Sup-CBOW & $3.5$ & $2344$ & $259$\\
    \cline{2-5}
    &\multicolumn{4}{c|}{\emph{models}}\\
    \cline{2-5}
    & N-Gram & $19.1$ & $285$ & $17$ \\
    & N-Gram w/cache & $19.1$ & $270$  & $18$  \\
    & RNN & $15.4$& $277$ &$24$\\
    & LSTM & $21.9$ & $149$ & $12$ \\
    & Memory Network & $8.5$ & $566$ & $46$ \\
    \hline
  \end{tabular}
  \caption{Results of computational methods. Accuracy is expressed in percentage.}
  \label{tab:results}
\end{table*}

\paragraph{Results} Results of models and baselines are reported in Table \ref{tab:results}. Note that the measure
of interest for LAMBADA is the average success of a model at
predicting the target word, i.e., accuracy (unlike in standard
language modeling, we know that the missing LAMBADA words can be
precisely predicted by humans, so good models should be able to
accomplish the same feat, rather than just assigning a high
probability to them). However, as we observe a bottoming effect with
accuracy, we also report perplexity and median rank of correct word,
to better compare the models.

As anticipated above, and in line with what we expected, all our
models have very good performance when called to perform a standard
language modeling task on the control set. Indeed, 3 of the models
(the N-Gram models and LSTM) can guess the right word in about 1/5 of
the cases.

The situation drastically changes if we look at the LAMBADA results,
where all models are performing very badly. Indeed, no model is even
able to compete with the simple heuristics of picking a random word
from the passage, and, especially, a random capitalized word (easily a
proper noun). At the same time, the low performance of the latter
heuristic in absolute terms (7\% accuracy) shows that, despite the
bias in favour of names in the passage, simply relying on this will
not suffice to obtain good performance on LAMBADA, and models should
rather pursue deeper forms of analysis of the broader context (the
Sup-CBOW baseline, attempting to directly exploit the passage in a
shallow way, performs very poorly).  This confirms again that the
difficulty of LAMBADA relies mainly on accounting for the information
available in a broader context and not on the task of predicting the
exact word missing.

In comparative terms (and focusing on perplexity and rank, given the uniformly low accuracy results) we
observe a stronger performance of the traditional N-Gram models over
the neural-network-based ones, possibly pointing to the difficulty of
tuning the latter properly. In particular, the best relative
performance on LAMBADA is achieved by N-Gram w/cache, which takes
passage statistics into account. While even this model is effectively
unable to guess the right word, it achieves a respectable
perplexity of 768.


We recognize, of course, that the evaluation we performed is very
preliminary, and it must only be taken as a proof-of-concept study of
the difficulty of LAMBADA. Better results might be obtained simply by
performing more extensive tuning, by adding more sophisticated
mechanisms such as attention \cite{Bahdanau:2014}, and so
forth. Still, we would be surprised if minor modifications of the
models we tested led to human-level performance on the task. %

We also note that, because of the way we have constructed LAMBADA,
standard language models are bound to fail on it by design: one of our
first filters (see Section \ref{sec:data-coll-meth}) was to choose
passages where a number of simple language models were failing to
predict the upcoming word. However, future research should find ways
around this inherent difficulty. After all, humans were still able to
solve this task, so a model that claims to have good language understanding
ability should be able to succeed on it as well.



\section{Conclusion}
\label{sec:conclusions}

This paper introduced the new LAMBADA dataset, aimed at testing language
models on their ability to take a broad discourse context into account
when predicting a word. A number of linguistic phenomena make the
target words in LAMBADA easy to guess by human subjects when they can
look at the whole passages they come from, but nearly impossible if
only the last sentence is considered. Our preliminary experiments
suggest that even some cutting-edge neural network approaches that are
in principle able to track long-distance effects are far from passing the LAMBADA challenge.

We hope the computational community will be stimulated to develop
novel language models that are genuinely capturing the non-local
phenomena that LAMBADA reflects. To promote research in this
direction, we plan to announce a public competition based on the
LAMBADA data.\footnote{The development set of LAMBADA, along with the training corpus, can be downloaded at \url{http://clic.cimec.unitn.it/lambada/}. The
  test set will be made available at the time of the competition.}

Our own hunch is that, despite the initially disappointing results of
the ``vanilla'' Memory Network we tested, the ability to store
information in a longer-term memory will be a crucial component of
successful models, coupled with the ability to perform some kind of
reasoning about what's stored in memory, in order to retrieve the
right information from it.

On a more general note, we believe that leveraging human performance
on word prediction is a very promising strategy to construct
benchmarks for computational models that are supposed to capture
various aspects of human text understanding. The influence of broad
context as explored by LAMBADA is only one example of this 
idea.

\section*{Acknowledgments}
We are grateful to Aurelie Herbelot, Tal Linzen, Nghia The Pham and, especially,
Roberto Zamparelli for ideas and feedback. This project has received
funding from the European Union's Horizon 2020 research and innovation
programme under the Marie Sklodowska-Curie grant agreement No 655577
(LOVe); ERC 2011 Starting Independent Research Grant n.~283554
(COMPOSES); NWO VIDI grant n.~276-89-008 (Asymmetry in Conversation).  
We gratefully acknowledge the support of NVIDIA Corporation with the donation 
of the GPUs used in our research.



\bibliography{../marco,other}
\bibliographystyle{acl2016}



\end{document}